\documentclass{ws-ijufks}
\usepackage[super]{cite}
\usepackage{placeins}

\usepackage[T1]{fontenc}
\usepackage{float}%
\usepackage{graphicx}%
\usepackage{multirow}%
\newcommand{\text}[1]{\mathrm{#1}}

\usepackage[title]{appendix}%
\usepackage{xcolor}%
\usepackage{textcomp}%
\usepackage{manyfoot}%
\usepackage{booktabs}%
\usepackage{listings}%
\usepackage{subcaption}

\usepackage{algorithm2e}

\usepackage{color}
\definecolor{orange}{RGB}{255,127,0}
\definecolor{black}{RGB}{0,0,0}

\definecolor{brown}{RGB}{150,70,0}
\definecolor{green}{RGB}{127,255,127}
\definecolor{mediumgreen}{RGB}{63,167,63}
\definecolor{darkgreen}{RGB}{0,127,0}
\definecolor{blue}{RGB}{127,127,255}
\definecolor{lightblue}{RGB}{150,150,255}
\definecolor{darkblue}{RGB}{0,0,127}
\definecolor{red}{RGB}{255,90,90}
\definecolor{grey}{RGB}{127,127,127}
\definecolor{pink}{RGB}{255,180,180}

\newcommand{\modified}[1]{{\textcolor{black} {#1}}}
\newcommand{\inserted}[1]{{\textcolor{black}{#1}}}

\begin{document}

\markboth{M. K\"angsepp et al.}{Calibrated Perception Uncertainty for Autonomous Driving}

%%%%%%%%%%%%%%%%%%%%% Publisher's Area please ignore %%%%%%%%%%%%%%%
%
\catchline{}{}{}{}{}
%
%%%%%%%%%%%%%%%%%%%%%%%%%%%%%%%%%%%%%%%%%%%%%%%%%%%%%%%%%%%%%%%%%%%%

%
\title{Calibrating Perception Uncertainty for Autonomous Driving  %% for Region Occupancy Queries in Bird's-Eye-View 
}
%
%\titlerunning{Calibrating Perception Uncertainty}
% If the paper title is too long for the running head, you can set
% an abbreviated paper title here
%

%\author{First Author\inst{1}\orcidID{0000-1111-2222-3333} \and
%Second Author\inst{2,3}\orcidID{1111-2222-3333-4444} \and
%Third Author\inst{3}\orcidID{2222--3333-4444-5555}}
%
%\authorrunning{F. Author et al.}
% First names are abbreviated in the running head.
% If there are more than two authors, 'et al.' is used.
%
%\institute{Princeton University, Princeton NJ 08544, USA \and
%Springer Heidelberg, Tiergartenstr. 17, 69121 Heidelberg, Germany
%\email{lncs@springer.com}\\
%\url{http://www.springer.com/gp/computer-science/lncs} \and
%ABC Institute, Rupert-Karls-University Heidelberg, Heidelberg, Germany\\
%\email{\{abc,lncs\}@uni-heidelberg.de}}
%

\author{Markus K\"angsepp}

\address{Institute of Computer Science, University of Tartu, \\
Narva mnt 19, Tartu 51009, Tartumaa, Estonia  \\
markus.kangsepp@ut.ee} 

\author{Meelis Kull}

\address{Institute of Computer Science, University of Tartu, \\
Narva mnt 19, Tartu 51009, Tartumaa, Estonia  \\
meelis.kull@ut.ee} 

%\authorrunning{M. K\"angsepp et al.}

\maketitle              % typeset the header of the contribution

\begin{history}
\received{(received date)}
\revised{(revised date)}
%\accepted{(Day Month Year)}
%\comby{(xxxxxxxxxx)}
\end{history}

\begin{abstract}

Autonomous driving systems must be capable of making quick decisions based on the perceived environment and specific driving conditions.
Perception models in these systems perform well in detecting objects under favourable conditions but their performance deteriorates in poor visibility or with partly occluded objects. 
To reduce risks from undetected objects, autonomous vehicles must incorporate all relevant uncertainties into their decision-making processes.
Grid-based perception outputs, such as occupancy grids, and object-based outputs, like lists of detected objects, must be accompanied by well-calibrated uncertainty estimates. 
These uncertainties are essential for ensuring the model's reliability and safety.

In this paper, we identify limitations in the current state-of-the-art and propose a more comprehensive set of uncertainty estimates that should be reported. 
In addition to commonly estimated forms of uncertainty about the presence, location, shape, and trajectory of detected objects, we propose to quantify the uncertainty about undetected objects within a region. 
Access to this set of uncertainties enables planners to perform region occupancy queries, which provide the probability that a certain region, such as the area around a chosen trajectory, is free of obstacles. 
We propose a novel approach for generating these probabilistic outputs from bird’s-eye-view (BEV) probabilistic semantic segmentation. 
Our experiments demonstrate that the initial probabilistic outputs from segmentation are not calibrated, and we present methods to achieve well-calibrated uncertainty estimates. 
Finally, we conduct an experiment on a downstream task, underscoring the importance of calibrated uncertainties for planning and highlighting the advantages of including additional uncertainty types.

\end{abstract}


\section{Introduction}
\label{sec:intro}  % related work'lt viide võetud ja pandud siia
%Introduction to problem, related topics, current state, research problems and questions, our approach.

Autonomous vehicles (AVs) are more popular and feasible than ever, relying heavily on various sensors such as cameras, LiDAR, radar, and GPS.
Given the high travel speeds of vehicles in traffic, the planning and decision-making subsystem of AVs must make rapid and justified decisions regarding driving speed and trajectory, while avoiding other vehicles, pedestrians, and obstacles~\cite{koopman2022how}.
To ensure safety, reliable and uncertainty-aware outputs are required from the perception subsystem responsible for detecting surrounding objects~\cite{veres2010,gruyer2017perception}.
Planning and decision-making methods for autonomous driving (AD) are increasingly able to utilise perception uncertainties~\cite{claussmann2019review}, aiding in safe and informed decision-making.

There are numerous perception methods for motion planning and for representing driving situations~\cite{claussmann2019review}.
We categorise the outputs of existing methods into two major types: \emph{grid-based view} and \emph{object-based view}.
The \emph{grid-based view} assigns a probability value to each cell in the grid, indicating the likelihood of an object being present in that specific cell. \inserted{A grid refers to a spatial map that divides the environment into uniformly sized cells. Depending on the application, these cells may correspond to pixels or to predefined physical areas (e.g., squares with side lengths of 0.5 meters).}
Grid-based approaches, such as occupancy grids~\cite{mouhagir2016integrating}, provide valuable information about each location but lack object-specific details, such as object class, size, and precise location, which are critical for behavioural modelling~\cite{claussmann2019review}.
To address this, it is necessary to explicitly detect individual objects, which we refer to as the \emph{object-based view}.
The \emph{object-based view} contains a list of detected objects, typically represented by bounding boxes that include information on object location and size.

%However, due to computational limitations, the list of detected objects in the \emph{object-based view} is non-exhaustive; there may be undetected objects, while the \emph{grid-based view} may still show a small probability (e.g., less than one-in-a-million) for the presence of something at that location.
Due to computational limitations, the list of detected objects in the \emph{object-based view} is typically non-exhaustive, in the sense of discarding the long tail of very unlikely but still plausible objects that fall below the detection threshold.
In contrast, the \emph{grid-based view} might still indicate a small probability (e.g., less than one-in-a-million) for the presence of an object at that location.
Even if the risk of having an object is very small, ignoring it can be hazardous, as the risk accumulates over time and across many grid cells.
%especially in adverse weather conditions, such as fog.
Thus, information from both grid-based and object-based views is complementary, and it is preferable for the perception system to provide both types of information.

The current approaches to representing perception with uncertainty have several limitations.
Firstly, they do not effectively combine \emph{grid-based} and \emph{object-based views} when representing uncertainty.
Furthermore, having uncertainty estimates alone is insufficient if these estimates are not well-calibrated. Well-calibrated uncertainty estimates would align predicted probabilities with real-world frequencies, ensuring that, for example, predicting a pedestrian with 80\% probability in 100 situations means that there should indeed be a pedestrian 80 times out of those 100 situations.

%Secondly, for safety, it is crucial to calculate the probability that the road ahead is safe to drive on — meaning there are no vehicles, pedestrians, or obstacles other than those detected by the perception system.
%Such probability is essential for determining a safe driving speed.
%However, to our knowledge, existing perception systems do not explicitly provide the probability of undetected objects in a region.
Secondly, for safety, it is essential to estimate the probability of undetected objects in the region ahead --- such as vehicles, pedestrians, or obstacles that the perception system may have missed. Understanding this probability is crucial for determining a safe driving speed, as it reflects the potential risks not accounted for by detected objects alone. However, to our knowledge, existing perception systems do not explicitly provide the probability of undetected objects in a region.

Lastly, planners often fail to fully utilise the information available from the perception system.
Some planners use object-based uncertainty~\cite{tang2022driving, lutjens2019safe}, while others optimise trajectories based on individual obstacle probabilities in each grid cell~\cite{artunedo2020motion, van2024overcoming}.
However, this assumes independence between uncertainty estimates across grid cells~\cite{ocallaghan2009Cont}, which, as we demonstrate, is strongly violated due to spatial auto-correlation (Figure~\ref{fig:autocor}).
Instead, we propose a novel approach called `region occupancy query' (ROQ), which planners can use as part of the interface to the perception system.
ROQ provides the probability that a given region is unoccupied, enabling planners to make safer, more reliable decisions regarding speed and trajectory.
%Additionally, ROQs can be used to assess the risk of collision. Typically, collision risk is calculated based on detected objects and their collision probabilities, without directly considering uncertainties.
%While some studies~\cite{lew2023risk, jasour2018moment, jasour2022fast, sharma2023hilbert, ledent2019formal, lee2019collision, van2024overcoming} use uncertainties for collision risk calculation, they do not account for the possibility of undetected objects or ensure the calibration of uncertainties.
Additionally, ROQs can be used as part of the information necessary to assess the risk of collision. 
While some studies~\cite{lew2023risk, jasour2018moment, jasour2022fast, sharma2023hilbert, ledent2019formal, lee2019collision, van2024overcoming} use uncertainties for collision risk calculation, they do not account for the possibility of having undetected objects.
Furthermore, these works do not investigate whether the uncertainty estimates are calibrated.

\subsection{Contributions}

The main contributions of this paper are as follows:

\begin{itemize}
    \item
    %We propose a novel methodology for uncertainty quantification in perception, combining grid-based and object-based uncertainty quantification methods, along with uncertainty calibration processes.
    We propose a new method for quantifying uncertainty in perception by combining grid-based methods, object-based methods, and uncertainty calibration.
    % We propose a new method for quantifying uncertainty in perception that builds upon probabilistic semantic segmentation outputs, combining grid-based and object-based methods with uncertainty calibration.
    \item
    %We introduce region occupancy queries, a novel addition to the planning interface, enabling planners to obtain additional probability information regarding whether a region is occupied or empty.
    We extend the interface between perception and planning by introducing region occupancy queries, allowing planners to obtain additional probabilistic information about whether a region is occupied or empty, thereby improving safety.
    \item
    %We argue that successful planning requires both grid-based and object-based uncertainty estimates to be present and calibrated. We demonstrate that performance in a downstream planning task decreases if these uncertainty estimates are not utilized.
    We argue that successful planning requires both grid-based and object-based uncertainty estimates to be present and calibrated. We demonstrate that in a downstream planning task, performance decreases when these uncertainty estimates are not utilised.
\end{itemize}
The out-of-distribution aspect of uncertainty quantification~\cite{amodei2016concrete} remains beyond the scope of this paper and is an important direction for future research.

\section{Background}
\label{sec:background}

This section provides an overview of how uncertainties have been extracted in previous studies, how collision risk can be calculated, and what uncertainties are typically used as inputs in planning.

\subsection{Model outputs in object-based and grid-based views}

In an object-based view, models typically output bounding boxes indicating detection locations and sizes. 
For instance, \textit{EfficientNet}~\cite{tan2019efficientnet} provides bounding boxes for each object within the camera frame. 
Gaussian YOLOv3~\cite{choi2019gaussian} provides a richer output by modelling uncertainties related to bounding box locations. 
%Other works ; 
Two papers by Feng et al.~\cite{feng2021labels, feng2019can} add uncertainty to object detections using LiDAR inputs.
\inserted{LiDAR (Light Detection and Ranging) is a sensing technology that emits laser beams and calculates distances by measuring the time required for the reflected signals to return, thereby enabling accurate mapping of the surrounding environment.} %propose approaches to incorporate uncertainty. 
\inserted{The first paper by Feng et al.~\cite{feng2021labels} introduces a generative model to determine the uncertainty in ground-truth bounding boxes and visualises it as spatial uncertainty in a bird's-eye view (BEV). }\inserted{A bird’s-eye view provides a top-down, map-like perspective of the environment, making it particularly useful for route planning.}
\inserted{The second study by Feng et al.~\cite{feng2019can} suggests methods to calibrate object detection uncertainty for a 3D LiDAR model.}

\begin{figure*}
    \begin{center}
    \includegraphics[width=0.7\textwidth]{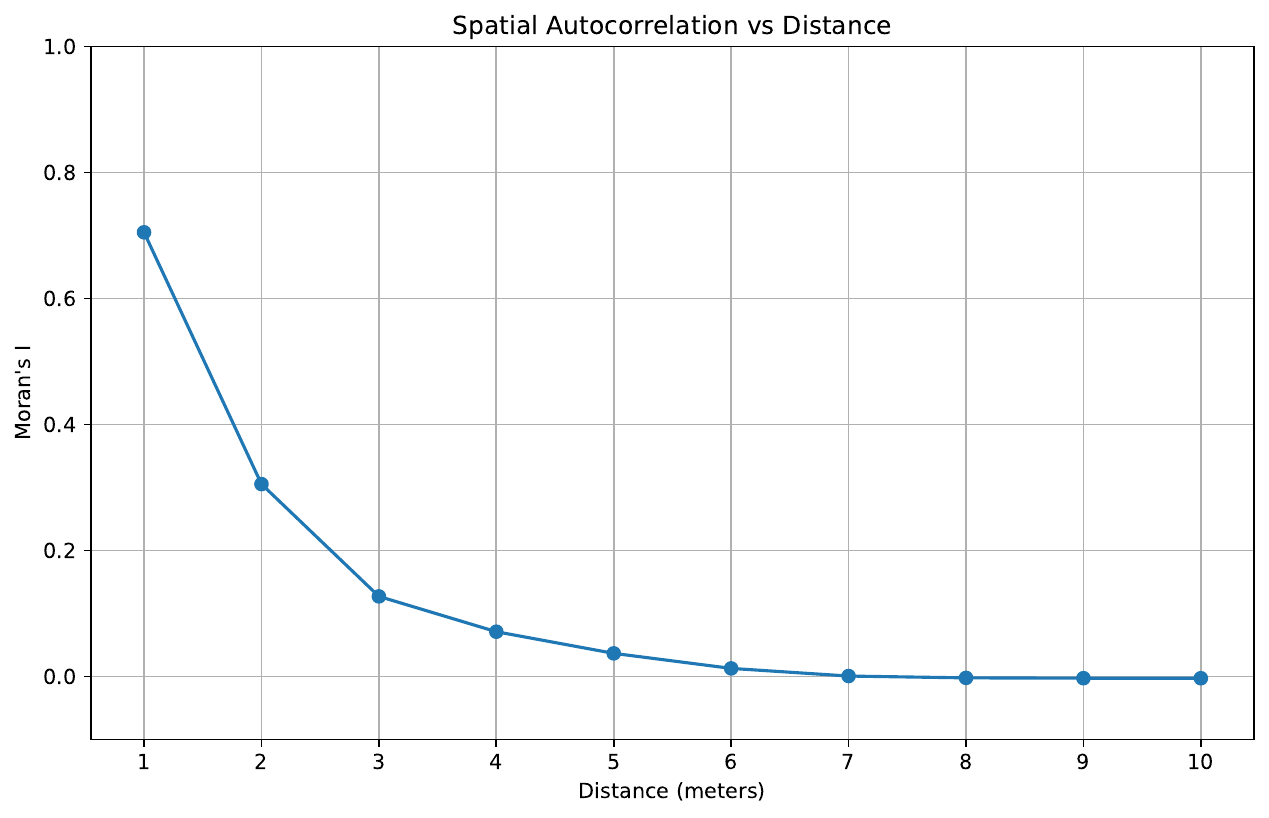}    
    \caption{
    %The correlation of grid cells by distance in meters. Each distance calculation excludes shorter distances, i.e., values between 1 and 2 meters are used for a distance of 2 meters. Distances are Euclidean and rounded to the nearest whole number. The figure is based on a BEV semantic segmentation from the FIERY model.
    Correlation between grid cells as a function of Euclidean distance in meters. For each distance $D$, correlations are computed using cell pairs separated by distances greater than $D - 1$  meters and up to $D$ meters. The figure is based on BEV semantic segmentation outputs from the FIERY model.
    }
    \label{fig:autocor}
    \end{center}
\end{figure*}

The grid-based view typically readily provides individual uncertainty estimates for each grid cell, e.g. in FIERY~\cite{hu2021fiery}, the BEV semantic segmentation is probabilistic.
\inserted{FIERY is a bird’s-eye-view semantic segmentation model that predicts vehicle positions at both the current and future timesteps. While classical semantic segmentation labels pixels in an image grid, FIERY performs segmentation over bird’s-eye-view grids in which each cell corresponds to a fixed region in the ground plane. For each grid cell, the model outputs the probability that it is occupied by an object. Within the grid-based view, this value can be regarded as an objectness score: higher scores indicate a higher probability of a true positive detection, while lower scores suggest a higher chance of a false positive.}
However, an important issue arises because the probabilities associated with nearby cells are not independent; there is spatial correlation between the probabilities of nearby cells~\cite{ocallaghan2009Cont}, as illustrated in Figure~\ref{fig:autocor}. A reason why this correlation exists is that nearby cells often contain information about the same object. As a result, the presence or absence of an object influences not just one cell but several cells in its vicinity. Consequently, cells that are closer together have higher correlation in their probabilities, while cells that are farther apart share less information about the same object, leading to decreased correlation between them.
%Despite this, many methods assume independence of occupancy maps. 

To some extent, an object-based view can be derived from multiple grid-based views, as demonstrated in the FIERY model~\cite{hu2021fiery}. For that, FIERY has an extra grid called centerness, which predicts object centre locations.

An uncertainty-aware approach is used in the grid-based view in works by Artunedo et al.~\cite{artunedo2020motion}, Yu et al.~\cite{yu2016semantic}, and Godoy et al.~\cite{godoy2021grid}.
Artunedo et al.~\cite{artunedo2020motion} consider localisation uncertainty in a probabilistic grid-based view for motion planning, combining outputs from mapping and perception systems.
Yu et al.~\cite{yu2016semantic} model the surrounding environment as lane grids, propagating ego-vehicle pose uncertainty across grid cells. \inserted{The ego-vehicle is the vehicle carrying the sensors and cameras and thus serves as the central frame of reference. All object detections, coordinate systems, and predictions are defined relative to its position and orientation.}
Finally, Godoy et al.~\cite{godoy2021grid} incorporate perception uncertainty from ego-vehicle perception system and other road users into grid-based occupancy maps.

While BEV semantic segmentation can also be considered a grid-based view, few methods extract uncertainty information for this type of segmentation~\cite{mukhoti2018semseg, phan2019semseg, gustafsson2020semseg}.
These works estimate epistemic and aleatoric uncertainties in semantic segmentation using ensembling, dropout, and/or Bayesian networks.
Mukhoti et al.~\cite{mukhoti2018semseg} propose metrics to evaluate Bayesian model uncertainty estimations and present a probabilistic version of a model that outputs pixel-wise uncertainties.

\subsection{Collision risk in object-based and grid-based views}

%To avoid collisions with other vehicles, pedestrians, and obstacles, autonomous vehicles must calculate collision risk.
%The probability of avoiding a collision can be calculated as follows: $$P(Z>0) = 1 - P(Z \leq 0) = 1 - \prod_{j=1}^{n_{\text{obj}}}(1-p_{overlap,j}),$$ where $Z$ is the negative distance to an object, $n_{\text{obj}}$ is the number of objects, and $p_{overlap}$ is the probability of overlap. This formula is based on Lew et al.~\cite{lew2023risk}. The overlap probability represents the area that a detected object covers within the planned trajectory.
To help reduce the likelihood of collisions with other vehicles, pedestrians, and obstacles, autonomous vehicles calculate collision risk as one of several strategies. The probability of avoiding a collision can be calculated as:
$$P_{\text{no collision}} = \prod_{j=1}^{n_{\text{obj}}} (1 - p_{\text{overlap},j}),$$
where  $n_{\text{obj}}$  is the number of detected objects, and  $p_{\text{overlap},j}$  represents the probability that object  $j$  overlaps with the vehicle’s planned trajectory.
This formula, based on Lew et al.~\cite{lew2023risk}, assumes that the overlap events for different objects are independent. 
By multiplying these probabilities of not overlapping with each detected object, autonomous vehicles can estimate the total probability of avoiding a collision.

Overlap calculations can also incorporate uncertainties, as explored in~\cite{lew2023risk, jasour2018moment, jasour2022fast, sharma2023hilbert}. 
Lew et al.~\cite{lew2023risk} propose a method that manages aleatoric and epistemic uncertainties of perception systems and keeps risk constraints for the whole trajectory. %For this, the sample average approximation method is used.
Jasour et al.~\cite{jasour2018moment} use a moment-sum-of-squares approach for non-convex safety constraints in uncertain environments, including uncertainties in obstacle location, size, and geometry. 
Jasour et al.~\cite{jasour2022fast} propose non-sampling methods for assessing trajectory risk in autonomous vehicles, using Gaussian and non-Gaussian mixture models for location uncertainties. 
Finally, Sharma et al.~\cite{sharma2023hilbert} use Reproducing Kernel Hilbert Space (RKHS) embeddings to generate probable trajectories based on obstacle distribution, enabling sampling from this distribution.

In a grid-based view, collision risk is calculated in works by Ledent et al.~\cite{ledent2019formal}, Lee et al.~\cite{lee2019collision}, and Van der Ploeg et al.~\cite{van2024overcoming}. 
Ledent et al.~\cite{ledent2019formal} validate collision risk estimation in a Conditional Monte Carlo Dense Occupancy Tracker (CMCDOT), a probabilistic perception system that outputs Bayesian probabilistic occupancy grids and time-to-collision probabilities. 
Lee et al.~\cite{lee2019collision} propose a Collision Avoidance/Mitigation System (CAMS) that uses Predictive Occupancy Maps (POMS) to assess risks for proposed trajectories. 
Van der Ploeg et al.~\cite{van2024overcoming} introduce an approach for multi-object trajectory assessment, taking into account objects of varying sizes, shapes, and potential occlusions.

\subsection{Planning in object-based and grid-based views}

The previous subsection covered uncertainties in model outputs and collision risk calculations. 
Here, we outline planning methods that incorporate these uncertainties.

Tang et al.~\cite{tang2022driving} introduce a planning method that accounts for uncertainty in object locations by using potential fields to model movement constraints, while also considering risks such as vehicle rollover and the physical dynamics of the vehicle.
L{\"u}tjens et al.~\cite{lutjens2019safe} present a reinforcement learning framework that uses dropout and bootstrapping to acquire location uncertainty information for detections.

Secondly, grid-based planners that consider uncertainty: Van der Ploeg et al.~\cite{van2024overcoming} develop a planner that uses risk fields from occupancy maps of different sizes and shapes, accommodating occluded objects. 
Artunedo et al.~\cite{artunedo2020motion} propose a planning method that incorporates occupancy grids, taking location uncertainty into account.

% Main contribution
\section{Uncertainty-aware perception for autonomous driving}
\label{sec:unc-aware_perception}

%This section provides an overview of our approach to uncertainty-aware perception for autonomous driving, outlining the potential needs and benefits of incorporating such an approach.
This section provides an overview of our approach to uncertainty-aware perception for autonomous driving, highlighting how safety requirements necessitate this approach and outlining the potential benefits of its incorporation.

\subsection{Uncertainty in the perception-planning stack}

Uncertainty plays an increasingly significant role in perception and planning; however, planning algorithms have not yet fully adapted to it (see Section~\ref{sec:background}).
We propose that perception methods should output uncertainties for both object-based and grid-based views, and planners should have the option to request additional information on specific regions of interest via region occupancy queries (ROQ).
The need for uncertainty in perception is emphasised by McAllister et al.~\cite{mcallister2017safety}, who argue that perception systems should include semantic, geometric, and localisation information about the scene.

According to our proposal, a perception model should be able to provide:
\begin{enumerate}
    \item \emph{A list of detected objects} (object-based view), with the following details for each object:
    \begin{itemize}
        \item \emph{Presence uncertainty}: the probability that the detection is a true positive;
        \item \emph{Location uncertainty}: a probability distribution over possible locations of the object's centre;
        \item \emph{Shape uncertainty}: a probability distribution over possible shapes and orientations of the object;
        \item \emph{Trajectory uncertainty}\footnote{Often part of a separate behavioural modelling subsystem, but included here as some perception models, like FIERY, provide trajectory uncertainty.}: a probability distribution over possible future \mbox{trajectories} of the object.
    \end{itemize}
    
    \item \emph{Occupancy maps} (grid-based view), with each cell providing information about a small area.
    
    %\item \emph{Region occupancy query} (ROQ): our proposed addition, where ROQs output the probability that specific regions or areas are empty.
    \item \emph{Region Occupancy Query} (ROQ): our proposed addition, which allows planners to query for the probability that specific regions or areas are empty, i.e. do not include any objects.
\end{enumerate}

\begin{figure*}[t]
    \begin{center}
    \begin{subfigure}{0.33\textwidth}
        \caption{}
        \includegraphics[width=\textwidth]{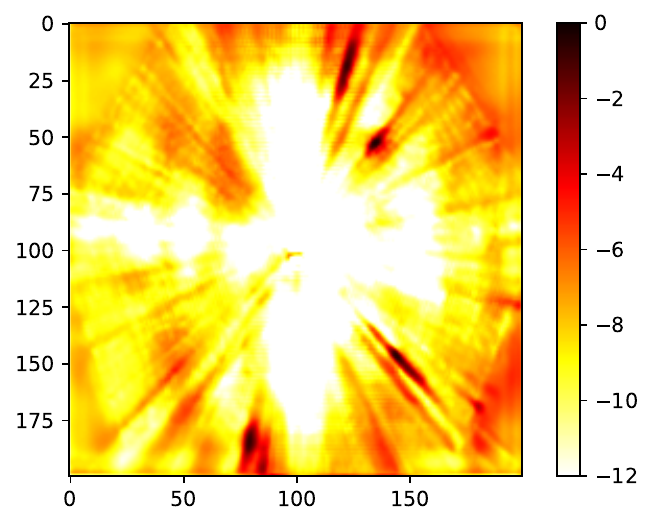}
        \label{fig:res_f1a1_outputs}
    \end{subfigure}%
    ~ %
    \begin{subfigure}{0.32\textwidth}
        \caption{}
        \includegraphics[width=\textwidth]{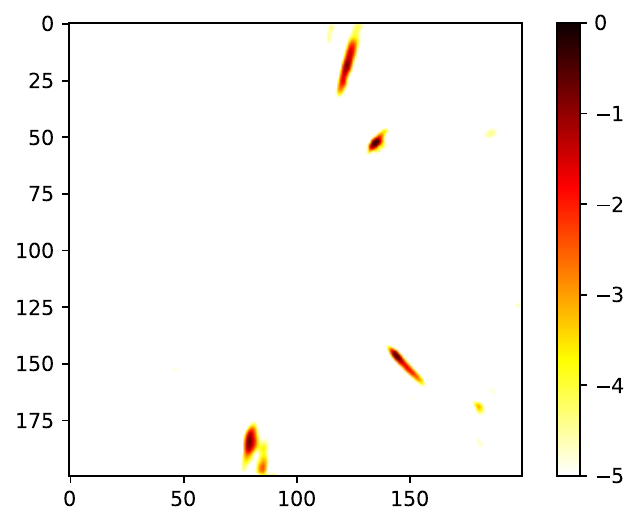}
        \label{fig:res_f1a2_outputs}
    \end{subfigure}%
    ~ %
    \begin{subfigure}{0.32\textwidth}
        \caption{}
        \includegraphics[width=\textwidth]{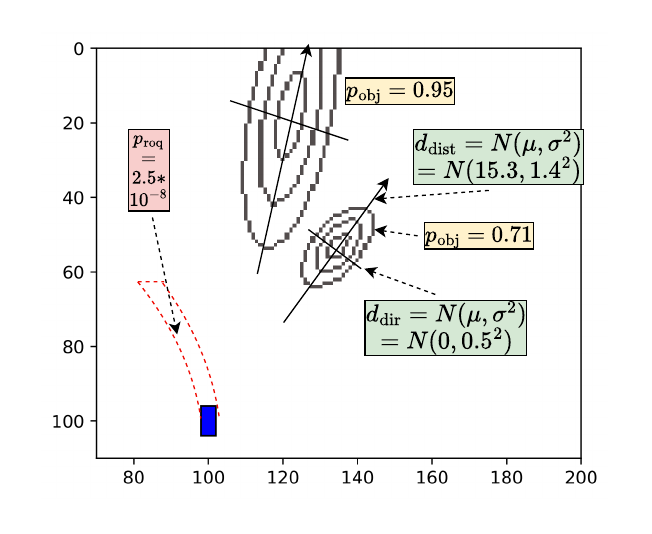}
        \label{fig:res_f1b_outputs}
    \end{subfigure}%
    \caption{
    %BEV perception uncertainty using the FIERY model. (a) and (b) show FIERY semantic segmentation in log-scale (natural logarithm) for improved visibility, clipped at -12 in (a) and -5 in (b). (c) displays our proposed additional information on object locations (ellipsoids) and an example ROQ region (dashed red shape). The blue rectangle at $(100,100)$ represents the ego-vehicle. Note that (c) is zoomed in. The example was generated using the method described in Section~\ref{sec:methods}.
    Visualisation of BEV perception uncertainty using the FIERY model. Figures (a) and (b) display the semantic segmentation outputs from FIERY, where colour intensities represent the natural logarithm of probabilities. In (a), log-probabilities are clipped at -12, and in (b), at -5. Figure (c) presents our proposed additional information on object locations depicted as ellipsoids, along with an example Region Occupancy Query (ROQ) region outlined by the dashed red shape. The blue rectangle at coordinates (100, 100) represents the ego-vehicle. Note that (c) is zoomed in for clarity. This example was generated using the method described in Section~\ref{sec:methods}.
    }
    \label{fig:segtoinputs}
    \end{center}
\end{figure*}

\begin{figure*}
    \begin{center}
    \includegraphics[width=1\textwidth]{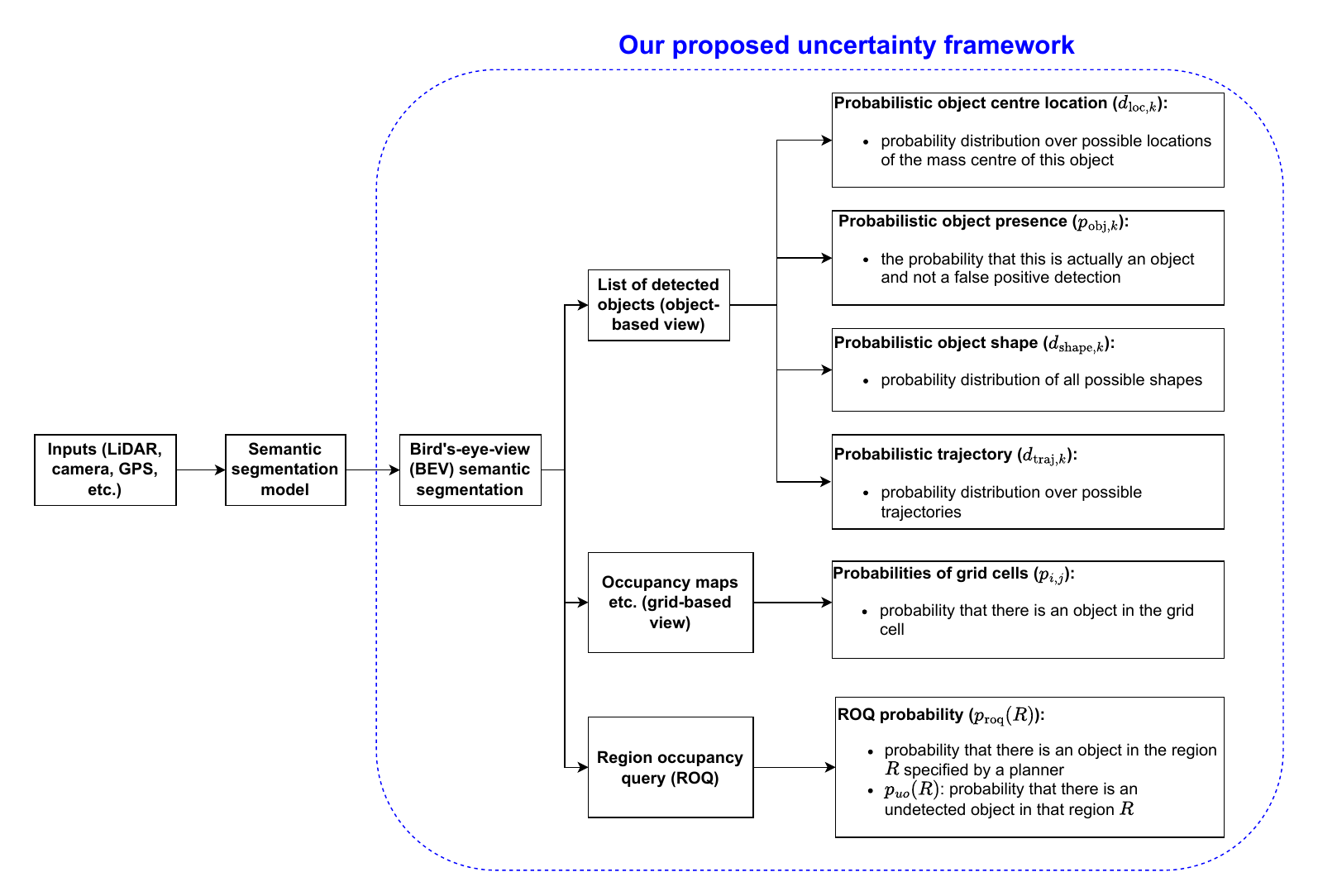}    
    \caption{Overview of our proposed uncertainty framework, showing the extracted uncertainties from BEV semantic segmentation. The blue dashed outline indicates the scope of our work.}
    \label{fig:framework}
    \end{center}
\end{figure*}

To clarify this proposal, consider the FIERY model trained to detect pedestrians (see Section~\ref{sec:experiments} for details). 
This model provides a bird's-eye-view (BEV) probabilistic semantic segmentation with 0.5m x 0.5m grid cells (see Figure~\ref{fig:res_f1a1_outputs}), where colors indicate log-scale probabilities of a grid cell containing an object, and the ego-vehicle is at $(100,100)$.
%\inserted{elaborate what is ego-vehicle, also in previous chapter ego-vehicle used, change that?}
Due to computational constraints in the perception-planning stack, the object-based view cannot include all positive signals from the grid-based view, as visualised by clipping at $-5$ in Figure~\ref{fig:res_f1a2_outputs}. Figure~\ref{fig:res_f1b_outputs} illustrates the location uncertainty of detected objects (ellipsoids) and an example region (red shape) ahead of the ego-vehicle (blue rectangle), for which our approach provides a ROQ probability. These regions can correspond to potential vehicle trajectories or specific areas, like pedestrian crossings. Figure~\ref{fig:framework} shows an overview of the proposed framework and its integration into the overall pipeline.

Below, we outline the need for each type of input to the planner and their advantages over planners that rely solely on segmentation or detected objects.

\textbf{The need for probabilistic object presence.}  
The planner must devise a trajectory that avoids all detected objects. 
However, if a 100\% safe trajectory is no longer feasible, it becomes crucial to identify which detected objects are more likely false positives.

\textbf{The need for probabilistic object centre location.}  
The planner relies on location uncertainties to estimate the object's position accurately and avoid collisions caused by localisation errors. 
High location uncertainty may indicate difficulty detecting objects in the current scenario.

\textbf{The need for probabilistic object shape.}  
Shape uncertainty accounts for possible variability in object dimensions and orientation.
It helps the planner estimate how much space the object might occupy.
High shape uncertainty may suggest occlusion or challenging conditions for shape prediction.

\textbf{The need for probabilistic object trajectory.}  
%This type of uncertainty assists the planner in predicting the future positions of dynamic objects and distinguishing static objects.
%High trajectory uncertainty prompts the planner to exercise additional caution around other road users.
This type of uncertainty provides the planner with predictions about the future positions of dynamic objects, differentiating them from static objects.
When trajectory uncertainty is high for certain objects, the planner must exercise extra caution specifically around these objects, as their future movements are less predictable.

\textbf{The need for occupancy maps.}  
%Occupancy maps (or other grid-based methods) enable planners to assess the likelihood of having an object covering any particular grid cell.
%This additional information is valuable for accounting for potentially undetected objects.
%However, due to the interdependence of grid cells (see Figure~\ref{fig:autocor}), planners may struggle to consider uncertainties across larger areas, underscoring the need for region occupancy queries.
Occupancy maps (or other grid-based methods) provide the planner with information about the likelihood that an object occupies each grid cell.
This data helps account for potentially undetected objects.
However, since uncertainties in neighbouring grid cells are correlated (see Figure~\ref{fig:autocor}), it can be difficult for the planner to evaluate uncertainty over larger areas, emphasizing the need for region occupancy queries.

\textbf{The need for region occupancy queries (ROQs).}  
ROQs help the planner assess the likelihood of some object occupying a particular region.
ROQs provide valuable information about the probability of missing objects in regions close to the ego-vehicle, aiding decisions on whether to avoid these areas or slow down due to uncertainty.

Each of these types of uncertainty estimates should be well-calibrated to accurately represent various driving scenarios and conditions, supporting reliable risk assessment and safe operation. %model validation.

\section{Methods}
\label{sec:methods}

This section presents our methodology for extracting the five proposed outputs (object presence, object centre location, object shape, object trajectory, and region occupancy query (ROQ)) from probabilistic BEV semantic segmentation. 
Additionally, we outline methods for calibrating these uncertainty estimates.

BEV semantic segmentation has been used in prior work as input for planners~\cite{paden2016survey}.
We believe that extending the set of perception outputs with well-calibrated uncertainties can significantly enhance safe operation of the perception-planning stack.

Our methods build upon any probabilistic BEV semantic segmentation model that outputs a matrix $\hat{P} = (\hat{p}_{ij})_{i,j=1}^{m,n}$, where $\hat{p}_{ij} \in [0,1]$ represents the predicted probability that an object overlaps with the grid cell $(i,j)$.
While we focus on a single object type (e.g., pedestrians or vehicles) and use the FIERY model~\cite{hu2021fiery} in our experiments (see Section~\ref{sec:experiments} for details), the methods are broadly applicable to any BEV probabilistic semantic segmentation model.

%For example, while the FIERY model~\cite{hu2021fiery} was originally trained to detect vehicles, we trained it to detect pedestrians.

The following subsections detail the methods for extracting uncertainty estimates from $\hat{P}$, followed by techniques for calibrating these estimates.
Initial extraction methods often rely on assumptions (e.g., Gaussian distributions, independence) that may not fully hold, resulting in uncalibrated uncertainty estimates. 
Calibration ensures that these uncertainties accurately represent the model's confidence in its predictions about the current situation.

\subsection{Extracting uncertainty estimates}
\label{sec:extract}

\textbf{The probabilistic object centre location ($d_{\text{loc}}$).} 
We model location uncertainty as a two-dimensional (2D) Gaussian in the BEV space. 
While this uncertainty is unlikely to be perfectly Gaussian, these estimates are calibrated later in the process. 
Notably, in our experiments, FIERY's segmentation predictions for pedestrians tend to exhibit ellipsoidal shapes (Figure~\ref{fig:res_f1a2_outputs}). 
Moreover, Gaussian Mixture Models (GMMs) are widely used for object location estimation in prior works~\cite{jasour2022fast, chai2019multipath, hong2019rules, deo2018multi}.
%We fit Gaussians for location uncertainty quantification using GMMs as follows. 
We quantify location uncertainty by fitting GMMs as follows.

%First, we select a suitable number of GMM components by identifying potential cluster centres based on semantic segmentation, using a probability threshold $p_{\text{thresh}}$. 
%For this, we adopt the same clustering method as used in the FIERY model for obtaining instance segmentation~\cite{hu2021fiery}.
%Next, we sample a large set of possible object locations with replacement, where the sampling probabilities are proportional to $\hat{p}_{ij}$ in individual grid cells.
%High-probability grid cells contribute more points, while low-probability grid cells contribute few or none. 
%We then fit a GMM to these sampled object locations. 

%To ensure proper stratification, we calculate the number of samples from each grid cell by multiplying the grid cell's probability $\hat{p}_{ij}$ by a constant parameter $m_{\text{thresh}}$ and rounding to the nearest integer.
%This count is expressed as $c_{xy}=\mathsf{round}(m_{\text{thresh}}\cdot\hat{p}_{ij})$, and each grid cell contributes $c_{xy}$ points to the dataset used for GMM fitting.
%Lowering $p_{\text{thresh}}$ and increasing $m_{\text{thresh}}$ allows for the detection of objects with lower probabilities, leading to more potential detections. 
%In our experiments, we set $m_{\text{thresh}}=\frac{1}{p_{\text{thresh}}}$, ensuring that a grid cell with a probability equal to the threshold contributes exactly one point.

First, we select a suitable number of GMM components by identifying potential cluster centres based on semantic segmentation. Specifically, we consider only the grid cells whose probabilities exceed a threshold $p_{\text{thresh}}$. These high-probability cells participate in the clustering process. We adopt the same clustering method used in the FIERY model for obtaining instance segmentation~\cite{hu2021fiery}.

Next, we sample object locations to fit the GMM. To ensure proper stratification, we calculate the number of samples from each grid cell by multiplying the grid cell’s probability $\hat{p}_{ij}$  by a scaling parameter  $m_{\text{thresh}}$  and rounding to the nearest integer:
$$
c_{ij} = \mathsf{round}(m_{\text{thresh}} \cdot \hat{p}_{ij}).
$$
Each grid cell contributes  $c_{ij}$ points to the dataset used for GMM fitting, hence high-probability cells contribute more points, while low-probability cells contribute few or none. In our experiments, we set  $m_{\text{thresh}} = \frac{1}{p_{\text{thresh}}}$, ensuring that a grid cell with a probability equal to the threshold contributes exactly one point. Lowering  $p_{\text{thresh}}$  and increasing  $m_{\text{thresh}}$  allows for the detection of objects with lower probabilities, leading to more potential detections.

Finally, we fit a GMM to these sampled object locations. The resulting GMM represents predicted object locations as a set of 2D Gaussians, each corresponding to an object detection. Algorithm~\ref{alg:loc} provides a structured overview of the process for determining probabilistic object centre locations.

%To summarize, the resulting GMM represents predicted object locations as a set of 2D Gaussians, each corresponding to an object detection.
%Algorithm~\ref{alg:loc} provides a structured overview of the process for determining probabilistic object centre locations.

\RestyleAlgo{ruled}
\SetKwComment{Comment}{/* }{ */}
\SetKwInput{kwInput}{Input}
\SetKwInput{kwCalculation}{Calculation}
\SetKwInput{kwReturn}{Return}

\begin{algorithm}
\caption{Probabilistic object centre location}\label{alg:loc}
\kwInput{
$(\hat{p}_{ij})_{i,j=1}^{m,n}$\;

$p_{\text{thresh}}$ \Comment*[r]{Threshold for potential cluster centres}
$m_{\text{thresh}} = \frac{1}{p_{\text{thresh}}}$\;
}

\kwCalculation{

Identify the number of potential cluster centres $n_{\text{nodes}}$ using $p_{\text{thresh}}$.

Fit a Gaussian Mixture Model (GMM) using $n_{\text{nodes}}$ and stratified sampling based on $c_{xy}$, where $c_{xy} = \mathsf{round}(m_{\text{thresh}} \cdot \hat{p}_{ij})$.
}
\kwReturn{$d_{\text{loc}}$ for $n_{\text{nodes}}$.}
\end{algorithm}

\textbf{The probabilistic object presence ($p_{\text{obj}}$).} 
For each detected object, the presence probability quantifies the likelihood (ranging from 0 to 1) that it represents an actual object rather than a false positive.
An object is considered a false positive if none of the grid cells overlapping its Gaussian ($d_{\text{loc},k}$) contain an object.
A grid cell is deemed to overlap with the Gaussian if it falls within the 99\% probability iso-contour of the multivariate Gaussian distribution.
%A grid cell is deemed to overlap with the Gaussian if it falls within 99\% of the probability mass.

To estimate $p_{\text{obj}}$, we sum the probabilities $\hat{p}_{ij}$ over the region covered by the Gaussian and divide by the expected number of grid cells ($S_{\text{exp}}$) typically covered by an object. 
In the experiments, we use $S_{\text{exp}}=5$ as the expected number of grid cells for pedestrians, typically consisting of one central cell and four adjacent cells.
In some cases, $p_{\text{obj}}$ may exceed 1 due to pixel-wise dependencies (see Figure~\ref{fig:autocor}). 
To address this, we clip $p_{\text{obj}}$ values to the range $[0,1]$.
Algorithm~\ref{alg:pres} provides a structured overview of the estimation process for probabilistic object presence.

\begin{algorithm}
\caption{Probabilistic object presence}\label{alg:pres}
\kwInput{
$(\hat{p}_{ij})_{i,j=1}^{m,n}$\;
$d_{\text{loc}}$ \Comment*[r]{Location distribution}
$S_{\text{exp}}$ \Comment*[r]{expected size of a typical object}
}

\kwCalculation{

Calculate $p_{\text{obj}} = \frac{\sum_{\hat{p}_{i,j \in d_{\text{loc},99\%}}} \hat{p}_{i,j}}{S_{\text{exp}}}$ \Comment*[r]{where $d_{\text{loc},99\%}$ means 99\% of Gaussian probability mass.}
Clip $p_{\text{obj}}$ to the range $[0,1]$.
}
\kwReturn{$p_{\text{obj}}$}
\end{algorithm}

%\textbf{The probabilistic object shape} is represented as a probability distribution  over all possible shapes. 
%Since BEV semantic segmentation provides object presence probabilities only per cell, estimating such a distribution reliably requires additional assumptions based on domain knowledge about the objects.

%If the model is sufficiently confident about the detected object's location, the location prediction can be used as a proxy for its shape.
%However, as the model's uncertainty about the location increases, the accuracy of shape estimation derived from semantic segmentation predictions decreases.
%In such cases, predefined shape distributions based on prior knowledge can be employed.

%In our experiments, we do not model shape uncertainty because our focus is on pedestrians, whose size in BEV space is relatively consistent compared to larger objects like vehicles.
%Furthermore, smaller objects may not be represented accurately in BEV grids with 0.5x0.5m cells, as these objects could overlap multiple cells, making them appear larger than they actually are.

\textbf{The probabilistic object shape ($d_{\text{shape}}$)} is represented as a probability distribution over all possible shapes. 
Since BEV semantic segmentation provides object presence probabilities only per cell, estimating such a distribution reliably requires additional assumptions based on domain knowledge about the objects.

In grid-based semantic segmentation, when the model assigns high confidence to the presence of an object in specific cells (pixels), the collection of these high-confidence cells can serve as an estimate of the object's shape.
However, this method assumes precise localisation and does not capture uncertainty in shape estimation effectively. 
As the model's confidence decreases or is spread across multiple cells, the accuracy of the shape derived from semantic segmentation predictions diminishes.
In such cases, predefined shape distributions based on prior knowledge can provide a more reliable estimate.

In our experiments, we do not model shape uncertainty because our focus is on pedestrians, whose size in BEV space is relatively consistent compared to larger objects like vehicles.
However, pedestrians may not be represented accurately in BEV grids with 0.5×0.5m cells, as they can overlap multiple cells, making them appear larger than they actually are.

\textbf{The probabilistic trajectory ($d_{\text{traj}}$)}  is modelled as the uncertainty in the speed and direction of the detected object or, alternatively, as uncertainty about the object's future locations.
Ideally, this information should be combined into a distribution of possible trajectories.

If the model does not directly provide trajectory information, it can be estimated by analysing the object's previous locations and timestamps.
This allows for the calculation of speed and direction, which can then be extrapolated to predict the object's movement in future timesteps.
Some variability should be added to account for potential changes in the object's course, as dynamic objects may not follow a consistent path.
The uncertainty in speed and direction is directly influenced by the location uncertainty in previous timesteps.

If the model does provide trajectory information, this data can be used directly.
Another approach is to leverage predictions of future semantic segmentations, as done by the FIERY model, instead of explicitly modelling the trajectory.

To summarise, the methods described above can be used to estimate location uncertainty for future predictions.

\textbf{The probabilities of region occupancy queries (ROQ) ($p_{\text{roq}}$).}
The goal is to estimate the probability of emptiness within a specific region of interest.

The probability $p_{\text{roq}}$ that region  $R$ is occupied can be calculated using the following formula:
% $$p_{\text{roq}}(R) = 1 - (1-p_{\text{uo}}(R))\prod_{k=1}^{n_{\text{obj}}}(1 - p_{\text{obj},k}\cdot\mathbb{E}_{d_{\text{loc},k},d_{\text{shape},k},d_{\text{traj},k}}[\hat{p}_{i,j \in R}]),$$
$$p_{\text{roq}}(R) = 1 - \left(1 - p_{\text{uo}}(R)\right) \prod_{k=1}^{n_{\text{obj}}} \left(1 - p_{\text{obj},k} \cdot p_{\text{overlap},k}(R) \right)$$
where $p_{\text{uo}}(R)$ is the probability of having undetected objects in this region, $n_{\text{obj}}$ is the number of detected objects, $p_{\text{obj},k}$ represents the probability of presence for object $k$, and $p_{\text{overlap},k}(R)$  is the probability that object  $k$  overlaps with region  $R$, accounting for uncertainties in its location, shape, and trajectory.
This formula computes the probability that region  $R$  is occupied by at least one object. It does so by considering both detected objects and the possibility of undetected objects. The term  $1 - p_{\text{uo}}(R)$  represents the probability that there are no undetected objects in  $R$. The product over all detected objects calculates the probability that none of the detected objects occupy region  $R$, accounting for their existence probabilities and uncertainties in their location, shape, and trajectory. Subtracting the combined probability of undetected and detected objects from $1$ yields the overall probability that the region is occupied.

To estimate $p_{\text{uo}}(R)$, the probability of undetected objects in region 
$R$, we sum the occupancy probabilities $\hat{p}_{i,j}$ for the individual grid cells within R $\setminus d_{\text{loc},99\%}$ --- that is, the grid cells in $R$ that are not overlapping with any detected objects (within their 99\% probability mass) --- and divide by the expected size $S_{\text{exp}}$ (in grid cells) of a typical object. 
Excluding cells overlapping with detected objects is essential because we are specifically interested in the probability of undetected objects occupying the region. This approach ensures that the uncertainties of detected objects and undetected regions can be calibrated separately.
Algorithm~\ref{alg:uo} provides a structured overview for calculating the probabilities of undetected objects.

%The probability of undetected objects ($p_{\text{uo}}(R)$) can be estimated similarly to the object's presence probability ($p_{\text{obj}}$).
%Specifically, this is achieved by summing up the probabilities $\hat{p}_{ij}$ for individual grid cells within the region $R$ and dividing by the expected size of an object.
%However, cells already overlapping with detected objects must be excluded, as the interest lies solely in the probability of undetected objects. 

%This exclusion is essential to ensure that the uncertainties of detected objects and regions of interest are calibrated separately.
%Note that $p_{\text{uo}}(R) \neq 1 - p_{\text{obj}}$, because $p_{\text{uo}}(R)$ relates to the probability of a region being empty of detected objects, whereas $1 - p_{\text{obj}}$ represents the likelihood that no object is present at a detected object's specific location.

%It is important to note that  $p_{\text{uo}}(R) \neq 1 - p_{\text{obj}}$  because  $p_{\text{uo}}(R)$  relates to the probability of undetected objects occupying region  $R$, whereas  $1 - p_{\text{obj},k}$  represents the probability that a specific detected object  $k$  is a false positive (i.e., does not actually exist at its predicted location).

By integrating both the uncertainties associated with detected objects and the possibility of undetected objects, this formula provides a comprehensive estimation of the occupancy probability for a given region. This estimation is crucial for autonomous vehicles to make informed decisions about navigating through or around uncertain areas, thereby enhancing safety and performance.

The ROQ formula incorporates all uncertainties proposed in this article. However, depending on the situation, certain uncertainties can be omitted, e.g. if insufficient information is available to derive reliable shape uncertainties for an object.

\begin{algorithm}
\caption{The probability of undetected objects in a region}\label{alg:uo}
\kwInput{
$(\hat{p}_{ij})_{i,j=1}^{m,n}$\;
$R$ \Comment*[r]{chosen region}
$d_{\text{loc}}$ \Comment*[r]{location distribution}
$S_{\text{exp}}$ \Comment*[r]{expected size of a typical object}
}

\kwCalculation{

$p_{\text{uo}}(R) = \frac{\sum_{\hat{p}_{i,j \in R \setminus d_{\text{loc},99\%}}} \hat{p}_{i,j}}{S_{\text{exp}}}$ \Comment*[r]{where $d_{\text{loc},99\%}$ means 99\% of Gaussian probability mass.}
Clip $p_{\text{uo}}(R)$ to the range $[0,1]$
}
\kwReturn{$p_{\text{uo}}(R)$}
\end{algorithm}

% \begin{algorithm}
% \caption{The probability of ROQ}\label{alg:roq}
% \kwInput{
% $R$ \Comment*[r]{chosen region}
% $n \gets n_{\text{obj}}$ \Comment*[r]{number of detected objects}
% $(p_{\text{obj},k})_{k=1}^{n}$ \Comment*[r]{probability of presence for each object}
% $(d_{\text{shape},k})_{k=1}^{n}$ \Comment*[r]{shape distributions}
% $(d_{\text{loc},k})_{k=1}^{n}$ \Comment*[r]{location distributions}
% $(d_{\text{traj},k})_{k=1}^{n}$ \Comment*[r]{trajectory distributions}
% $p_{\text{uo}}(R)$ \Comment*[r]{probability of undetected objects in region $R$}
% }

% \kwCalculation{
% $p_{\text{roq}}(R) = 1 - (1-p_{\text{uo}}(R))\prod_{k=1}^{n_{\text{obj}}}(1 - p_{\text{obj},k}*\mathbb{E}_{d_{\text{loc},k},d_{\text{shape},k},d_{\text{traj},k}}(\hat{p}_{i,j \in R}))$
% }
% \kwReturn{$p_{\text{roq}}(R)$}
% \end{algorithm}

\subsection{Uncertainty calibration}
\label{sec:calibration}

%As demonstrated in Section~\ref{sec:experiments}, the above uncertainty estimates are initially uncalibrated and require calibration. 
%First, the semantic segmentation probabilities $\hat{p}_{ij}$ can be calibrated.
%However, even after extracting object-based uncertainties from calibrated semantic segmentation, the results may still lack calibration.
%Therefore, we propose methods to calibrate pixel-based, object-based, and region-based uncertainty estimates. 
As demonstrated in Section~\ref{sec:experiments}, the initial uncertainty estimates are uncalibrated and require calibration. To address this, we can first calibrate the semantic segmentation probabilities $\hat{p}_{ij}$. However, after extracting object-based uncertainties from the calibrated semantic segmentation, these results may still lack proper calibration. Therefore, in the following we propose methods to calibrate uncertainty estimates at all levels: pixel-based, object-based, and region-based.

The goal of this work is not to identify the best calibration method for each type of uncertainty but to highlight the issue of miscalibration and show that simple calibration methods can effectively address it.

\textbf{Calibration of semantic segmentation.}  
Calibration of semantic segmentation can be performed pixel-wise, i.e. the same calibration map is applied independently on each pixel (whereby a pixel in the semantic output corresponds to a grid cell in BEV). We use isotonic regression~\cite{zadrozny2002transforming} on a holdout validation set, as it is efficient for large datasets. However, isotonic regression struggles to reliably predict very low probabilities due to the limited size of the validation set~\cite{allikivi2019non}. To address this, we clip the isotonic regression outputs at a minimum value of $\frac{1}{n_{\text{val}}}$, where $n_{\text{val}}$ is the number of validation instances.

\textbf{Calibration of location uncertainty.}  
Location uncertainty, represented as a 2D Gaussian, reflects the expected errors in estimating an object's location in both distance and direction. Our experiments reveal that the largest errors occur in estimating the distance from the ego-vehicle, while directional errors are relatively smaller. To simplify the calibration process, we approximate the 2D Gaussian as a product of two independent orthogonal 1D Gaussians: one for distance uncertainty and one for directional uncertainty. 

We calibrate the distance uncertainty and directional uncertainty separately using the quantile remapping method proposed by Kuleshov et al.~\cite{kuleshov2018accurate}. 
This method adjusts the predicted quantiles to achieve better calibration. 
We learn these calibration maps using all true positive predictions in the validation set.

\textbf{Calibration of presence uncertainty.}  
Presence uncertainty calibration is treated as calibration in binary classification.
We employ beta calibration~\cite{kull2017beta}, which is well-suited for smaller datasets due to its simplicity and reliance on only three parameters. 
Calibration maps are learned from all detected objects in the validation set. 
An object is labelled as present if 99\% of the probability mass in its location distribution overlaps with a grid cell containing an object.

\textbf{Calibration of undetected object probabilities in a region.}  
%Calibrating the probability of undetected objects in a region is another binary classification task, and we again use beta calibration. 
%Calibration maps are learned using all frames in the validation set. 
%However, only detections where the region does not overlap with any 99\% Gaussian location uncertainty region of a detected object are included.
%This ensures that the calibration focuses exclusively on regions without detected objects.
Calibrating the probability of undetected objects within a region is also a binary classification task, for which we again use beta calibration. We train the calibration maps using all frames in the validation set but include only regions that do not overlap with any detected object's 99\% Gaussian location uncertainty area. This ensures that the calibration focuses exclusively on areas without detected objects.

% Experiments
\section{Experiments}
\label{sec:experiments}

All experiments use the FIERY~\cite{hu2021fiery} BEV semantic segmentation model, which we have retrained on the pedestrian class of the NuScenes~\cite{nuscenes2019} dataset. 
%In contrast to the original study, which focused on vehicles, we selected the harder task of detecting pedestrians to better evaluate the calibration of uncertainty estimates, given the limited size of the test set. 
Unlike the original study that focused on vehicles, we chose the more challenging task of detecting pedestrians to better evaluate the calibration of uncertainty estimates, given the limited size of the test set.
Otherwise, we adopted the same experimental setup as the original FIERY study. 

The FIERY model predicts semantic segmentation with a resolution of 200x200 pixels, where each pixel corresponds to a 0.5x0.5 meter area, covering a range of 50 meters in all directions around the ego-vehicle.
FIERY uses data from past and current time steps to predict semantic segmentation for both current and future time steps.
To produce instance segmentation with trajectories, FIERY outputs four components: semantic segmentation, offset, flow, and centerness.
By combining these outputs with thresholding and clustering, it is possible to extract object-based information such as instance segmentation.
This enables differentiation between individual vehicles or pedestrians, unlike standard segmentation, which assigns only a class label (e.g., vehicle, pedestrian) to each grid cell.

The NuScenes dataset~\cite{nuscenes2019} comprises real-world driving sequences under diverse conditions in Boston and Singapore. 
It contains 700 labelled training sequences and 150 labelled validation sequences, each spanning approximately 30 frames.
For this study, the 150 validation sequences were split into 30 sequences for validation and 120 for testing.
Frames are captured at a rate of two per second, with each sequence covering about 20 seconds.

The experiments are divided into two main parts: calibration of semantic segmentation and calibration of detected and undetected objects.
Calibration is performed on approximately 20\% of the frames, while testing is conducted on the remaining 80\%.
To ensure unbiased evaluation, no driving episode in the test set has frames overlapping with the training or validation sets.
For simplicity, shape uncertainty is not modelled in these experiments; instead, each predicted pedestrian is assumed to occupy a fixed area of 5 pixels, which is the median size observed in the training set.

\subsection{Evaluation metrics and methods}

For evaluation of calibration, we use the expected calibration error (ECE)~\cite{naeini2015obtaining} and reliability diagrams~\cite{niculescu2005predicting}, which are standard in this domain. 

\begin{itemize}
    \item \emph{ECE} measures the weighted average difference between predicted probabilities and observed frequencies across equal-width or equal-size bins of confidence values. It quantifies the degree of miscalibration.
    \item \emph{Reliability diagrams} visualise the same information by plotting predicted probabilities on the x-axis and observed frequencies on the y-axis. The distance of bins from the diagonal indicates the deviation between predicted and calibrated probabilities (Figure~\ref{fig:reldiag_presence}).
\end{itemize}

To improve the readability of reliability diagrams, we follow the method proposed by Kängsepp et al.~\cite{kangsepp2022usefulness}.
This replaces the standard flat-topped bars with tilted-roof bars, allowing the distance from the diagonal to be visually assessed along the entire bar, not just at the centre marked by the red dot.
No information is lost, as the original bar height remains marked by the red dot.

\subsection{Comparison of uncertainty quantification methods}

The experiments compare three approaches for uncertainty quantification:
\begin{itemize}
    \item \emph{Uncalibrated (uncal)}: Direct extraction of object-wise uncertainties from uncalibrated semantic segmentation.
    \item \emph{Pixel-wise calibrated (pw-cal)}: Pixel-wise calibration of semantic segmentation followed by direct extraction of object-wise uncertainties.
    \item \emph{Object-wise calibrated (obj-cal)}: Calibration applied to the extracted object-wise uncertainties from uncalibrated semantic segmentation.
\end{itemize}

These methods allow us to assess the impact of calibration at different stages of the perception pipeline and demonstrate the importance of calibration for reliable uncertainty estimates.

\subsection{Probabilistic object locations}
\label{subsec:objloc}

For predicting object locations using a Gaussian Mixture Model (GMM), we set $p_{\text{thresh}} = 0.01$ and $m_{\text{thresh}} = 100$. The validation dataset includes 2,153 detections, while the test dataset contains 10,099 detections.

\begin{figure*}
    \begin{center}
    \begin{subfigure}{0.33\textwidth}
        \caption{}
        \includegraphics[width=\textwidth]{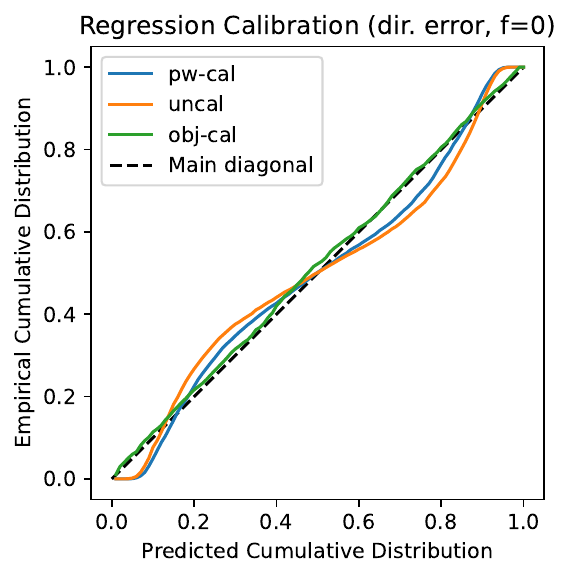}
        \label{fig:cal_location_width}
    \end{subfigure}%
    ~ %
    \begin{subfigure}{0.33\textwidth}
        \caption{}
        \includegraphics[width=\textwidth]{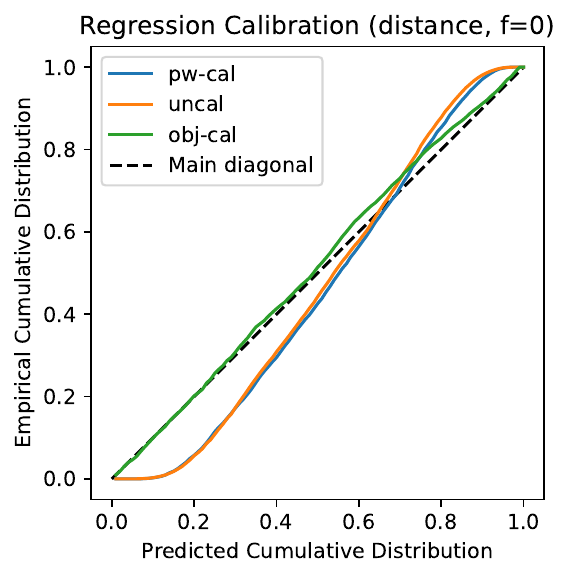}
        \label{fig:cal_location_distance}
    \end{subfigure}%
    \caption{Regression calibration plots~\protect\cite{kuleshov2018accurate} for object location probabilities: (a) direction error and (b) distance. Results are shown for uncalibrated (uncal), pixel-wise calibrated (pw-cal), and object-wise calibrated (obj-cal) cases. The black dashed line represents perfect calibration.}
    \label{fig:reldiag_location}
    \end{center}
\end{figure*}

Figures~\ref{fig:cal_location_width} and~\ref{fig:cal_location_distance} show regression calibration plots for location probabilities. 
These plots reveal that uncalibrated and pixel-wise calibrated cases deviate significantly from the diagonal, indicating poor calibration. 
In contrast, object-wise calibration achieves much closer alignment with the diagonal, demonstrating well-calibrated results.

For example, in the uncalibrated case, both ends of the curve are flat, suggesting an overabundance of predictions near zero and one.
This indicates that the model is overconfident in some predictions.
Additionally, for direction errors, the model exhibits both overconfidence and underconfidence, leading to a mix of calibration issues.

\subsection{Probabilistic object presence}
\label{subsec:objpres}

Probabilistic object presence is calibrated on the validation dataset containing 3,066 instances using beta calibration.
The test dataset consists of 14,214 instances.

\begin{figure*}
    \begin{center}
    \begin{subfigure}{0.33\textwidth}
        \caption{}
        \includegraphics[width=\textwidth]{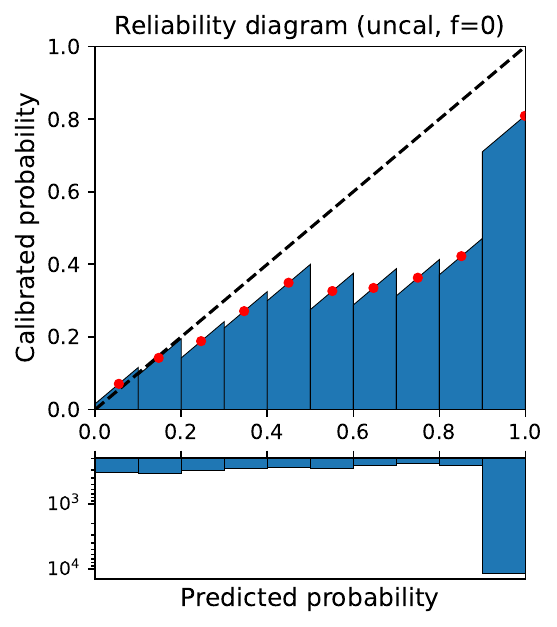}
        \label{fig:rel_diag_uncal_f0}
    \end{subfigure}%
    ~ %
    \begin{subfigure}{0.33\textwidth}
        \caption{}
        \includegraphics[width=\textwidth]{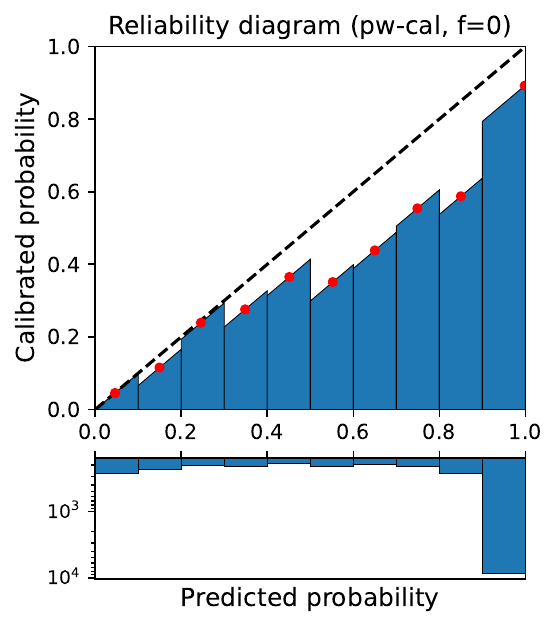}
        \label{fig:rel_diag_gmm_cal_f0}
    \end{subfigure}%
        ~ %
    \begin{subfigure}{0.33\textwidth}
        \caption{}
        \includegraphics[width=\textwidth]{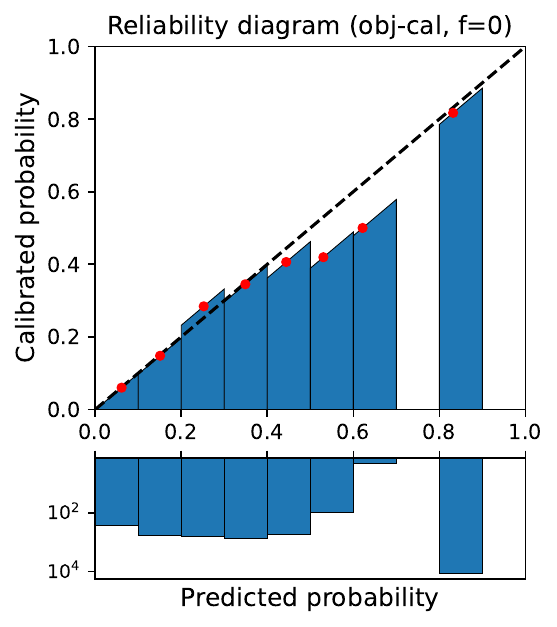}
        \label{fig:rel_diag_obj_cal_f0}
    \end{subfigure}%
    
    \caption{Reliability diagrams for object presence probabilities: (a) uncalibrated, (b) pixel-wise calibrated, and (c) object-wise calibrated. Histograms below the diagrams show the number of elements in each bin, with the y-axis log-scaled for better comparison. Instances are divided into 10 bins.}
    \label{fig:reldiag_presence}
    \end{center}
\end{figure*}

Figure~\ref{fig:reldiag_presence} shows reliability diagrams illustrating that uncalibrated predictions tend to overestimate the probability of pedestrian presence.
After calibration, these effects are significantly reduced, as reflected in the expected calibration error (ECE) scores: 0.185 (uncalibrated), 0.109 (pixel-wise calibrated), and 0.016 (object-wise calibrated).

Histograms beneath the reliability diagrams indicate that most uncalibrated predictions cluster at very high probabilities. 
Object-wise calibration redistributes these probabilities more evenly, improving the histogram balance and reducing overconfidence.

\subsection{Probabilistic trajectories of objects}

Object trajectories are represented as future locations $d_{\text{loc}}^{f=4}$ and the presence of objects $p_{\text{obj}}^{f=4}$, which correspond to predictions 2 seconds (4 frames) into the future. 

\modified{Figures~\ref{fig:rel_diag_uncal_f4}--\ref{fig:cal_location_distance_f4}} (see \ref{app:extra_exp}) indicate that the calibration performance for future time steps ($f=4$) is consistent with that for the current time step ($f=0$).
These results suggest that the proposed calibration method generalises well to future predictions, demonstrating the model's ability to accurately predict pedestrian trajectories over time.

\subsection{Probability estimation of undetected objects by regions}
\label{subsec:uoa}

As an example, our experiment focuses on a safety-critical region of 10x20 meters in front of the ego-vehicle to estimate the probability of missing an object. 

\begin{figure*}
    \begin{center}
    \begin{subfigure}{0.33\textwidth}
        \caption{}
        \includegraphics[width=\textwidth]{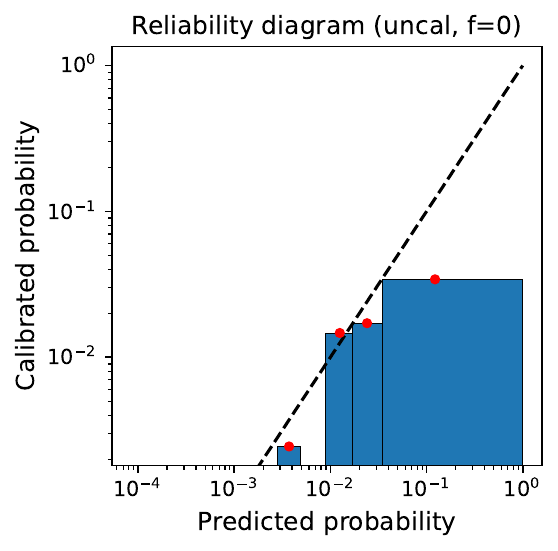}
        \label{fig:rel_diag_uncal_f0_undet}
    \end{subfigure}%
    ~ %
    \begin{subfigure}{0.33\textwidth}
        \caption{}
        \includegraphics[width=\textwidth]{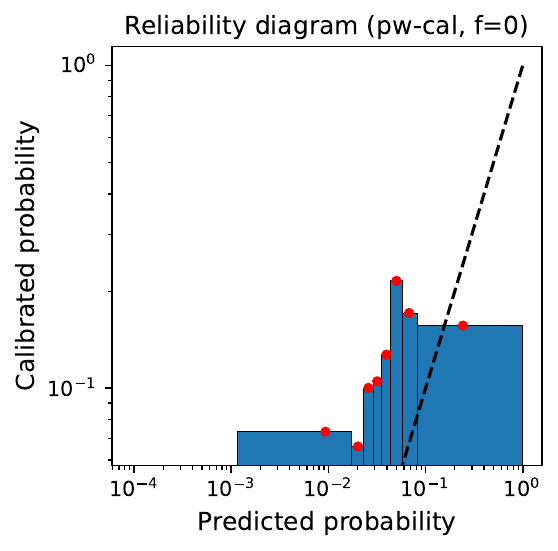}
        \label{fig:rel_diag_gmm_cal_f0_undet}
    \end{subfigure}%
    ~ %
    \begin{subfigure}{0.33\textwidth}
        \caption{}
        \includegraphics[width=\textwidth]{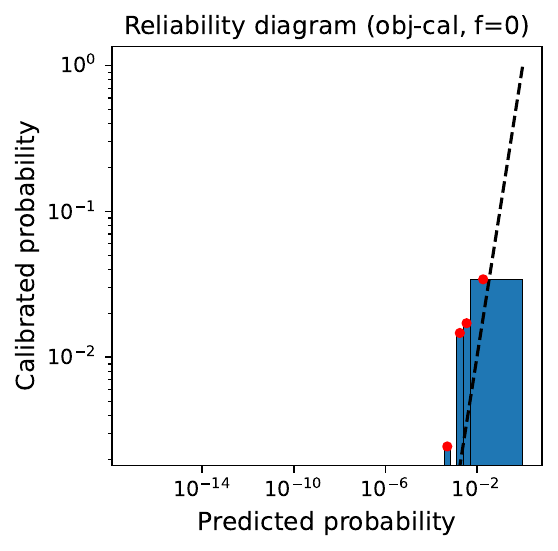}
        \label{fig:rel_diag_obj_cal_f0_undet}
    \end{subfigure}%
    
    \caption{Reliability diagrams for undetected object probabilities in the region: (a) uncalibrated, (b) pixel-wise calibrated, and (c) object-wise calibrated. Data is divided into equal-size bins, and axes are log-scaled for better visibility. Instances are divided into 10 bins.}
    \label{fig:reldiag_undetected}
    \end{center}
\end{figure*}

Figure~\ref{fig:reldiag_undetected} shows that the uncalibrated model underestimates probabilities in most bins, meaning the chance of missing an object is higher than predicted.
Pixel-wise calibration often overestimates probabilities, resulting in worse ECE scores. 
Object-wise calibration achieves the best results, with ECE scores of 0.011 (uncalibrated), 0.070 (pixel-wise calibrated), and 0.005 (object-wise calibrated).

Given the limited number of undetected objects (28 instances across 4,091 test frames), these results are indicative but not definitive. 
Additional data and regions should be considered for further analysis.

\subsection{Calibration of semantic segmentation}  
\label{subsec:calsegm}

Semantic segmentation calibration was performed on 20\% of validation frames (41 million pixels) and evaluated on the remaining 80\% of test frames (164 million pixels). Pixel-wise calibration significantly improved the expected calibration error (ECE) from 0.00038 (uncalibrated) to 0.000032, and the negative log-likelihood (NLL) from 0.00356 to 0.00329.

\begin{figure*}
    \begin{center}
    \begin{subfigure}[t]{0.32\textwidth}
        \caption{}
        \includegraphics[width=\textwidth]{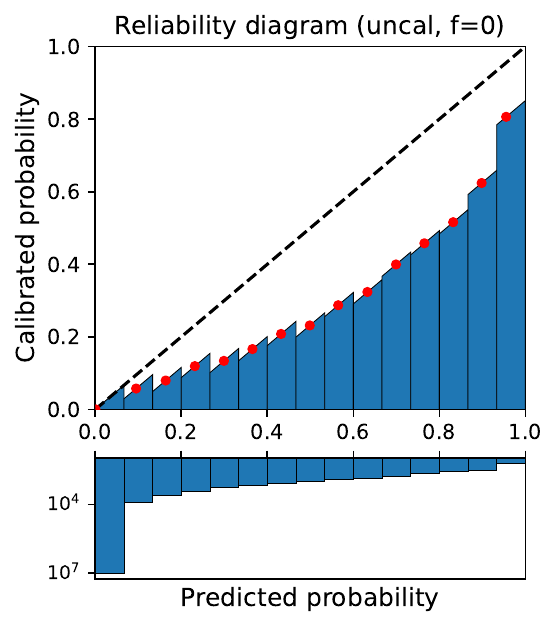}
        \label{fig:rel_diag_uncal_ew_f0_pw}
    \end{subfigure}%
    ~ %
    \begin{subfigure}[t]{0.32\textwidth}
        \caption{}
        \includegraphics[width=\textwidth]{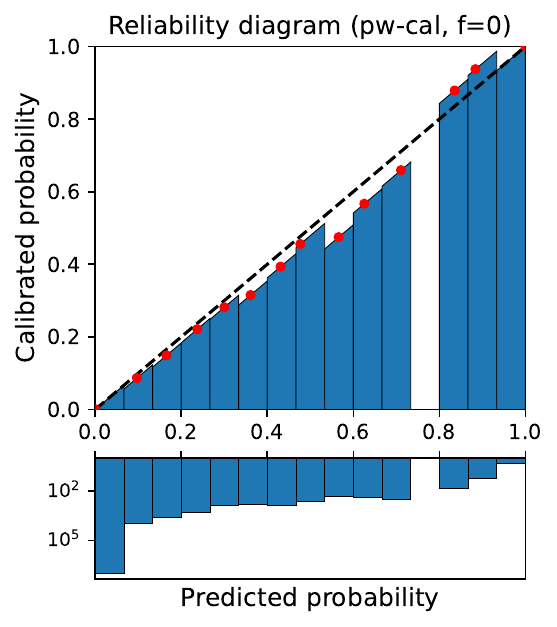}
        \label{fig:rel_diag_pwcal_ew_f0_iso}
    \end{subfigure}
    ~ %
    \begin{subfigure}[t]{0.31\textwidth}
        \caption{}
        \includegraphics[width=\textwidth]{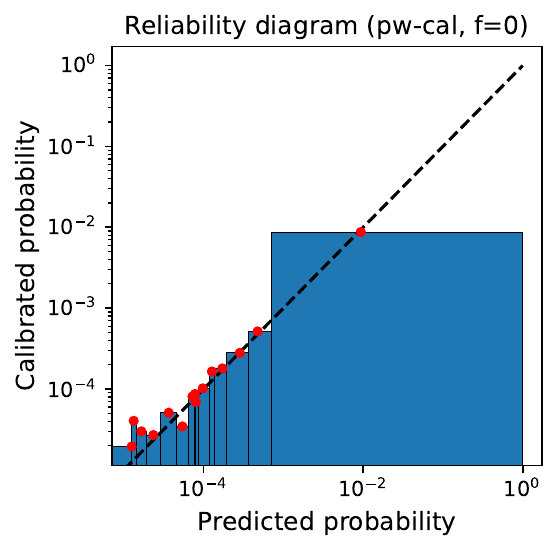}
        \label{fig:rel_diag_pwcal_es_f0_iso}
    \end{subfigure}%
    
    \caption{Reliability diagrams for pixel-wise semantic segmentation calibration: (a) uncalibrated with equal-width binning, (b) pixel-wise calibrated with equal-width binning, and (c) pixel-wise calibrated with equal-size binning. Instances are divided into 15 bins.}
    \label{fig:reldiag_pw_cal}
    \end{center}
\end{figure*}

Figure~\ref{fig:reldiag_pw_cal} illustrates that uncalibrated predictions tend to overestimate probabilities, as seen in Figure~\ref{fig:rel_diag_uncal_ew_f0_pw}.
After calibration, bins are closer to the diagonal, as shown in Figure~\ref{fig:rel_diag_pwcal_ew_f0_iso}.
Figure~\ref{fig:rel_diag_pwcal_es_f0_iso} demonstrates equal-size binning for low-probability pixels, confirming that even low-probability predictions are well-calibrated.

\subsection{Downstream task: trajectory planning}  
\label{subsec:planning}

In this subsection, we evaluate the proposed uncertainty quantification methods by applying them to a downstream trajectory planning task. 
The aim is to identify the best trajectory from a set of 25 predefined trajectories and to assess whether incorporating calibrated object location, presence, and undetected object uncertainties improves trajectory selection.
While this task is a simplification of the challenges in autonomous driving, it provides insights into the practical utility of the proposed methods.

The experiments use the NuScenes dataset, with 4,091 frames for testing and 1,027 validation frames for calibration.
For each frame, 25 predefined trajectories are used to determine the optimal trajectory.
Figure~\ref{fig:predefined_trajectories} (see~\ref{app:extra_exp}) visualises these trajectories.
Each trajectory represents the ego-vehicle's path over the next four timesteps (2 seconds), with a constant speed of 10 m/s.
The turning speed for each trajectory is also fixed, ranging from -0.3 to 0.3 radians per timestep.

The proposed methods are used to estimate the probability that a given trajectory avoids collisions with pedestrians.
For \emph{uncal} and \emph{pw-cal}, this probability is calculated by summing the semantic segmentation probabilities $\hat{p}_{ij}$ overlapping the trajectory. 
For \emph{ROQ (obj-cal)}, the probability combines the undetected object probability in the region with the presence probabilities of detected objects overlapping the trajectory, weighted by their overlap proportion.

These probabilities are then evaluated against the binary event indicators of whether a collision with a pedestrian occurred along the trajectory.
Since each trajectory spans four timesteps, the pedestrian locations at each timestep are taken into account.

\begin{table}
    \centering
    \caption{Comparison of trajectory estimation quality across three approaches: uncalibrated (uncal), pixel-wise calibrated (pw-cal), and object-calibrated with undetected-object-ahead probabilities (ROQ (obj-cal)). The best scores are highlighted in bold.}
    \begin{tabular}{lrrr}
    \hline
    {} &       NLL & BS & AUC\\
    \hline
    uncal      &  0.605926 &  0.034923 &  0.895200 \\ 
    pw-cal     &  0.536661 &  0.033925 &  0.903931 \\ 
    ROQ (obj-cal)    &  \textbf{0.269738} &  \textbf{0.032463} &  \textbf{0.904931} \\  
    \hline
    \end{tabular}
    \label{tab:planning}
\end{table}

To evaluate prediction performance, we use negative log-likelihood (NLL), Brier score (BS), and area under the ROC curve (AUC).
The focus is on whether the proposed calibration steps enhance planning performance.

A key objective is ensuring that trajectories with collisions are ranked lower than those without collisions.
AUC measures how effectively the uncertainty estimation methods rank the 25 trajectories.
As shown in Table~\ref{tab:planning}, the average AUC improves from \emph{uncal} to \emph{pw-cal} and further to \emph{ROQ (obj-cal)}.
This demonstrates that better-calibrated uncertainties enable more accurate ranking of trajectories.

While ranking helps select the best trajectory, managing overall risk and determining appropriate speeds also require accurate collision probability estimates.
These are evaluated using NLL and BS, which quantify the quality of probabilistic predictions.
Table~\ref{tab:planning} shows that object-wise calibration (\emph{ROQ (obj-cal)}) significantly improves both NLL and BS, confirming the importance of proper calibration for trajectory planning.

Runtime comparisons for these methods are provided in~\ref{sup:runtime}.

\section{Conclusion}

This paper introduced a novel approach to modelling uncertainties in object detection. We proposed methods to estimate uncertainties for object location, shape, presence, trajectory, and the probability of undetected objects in critical regions near the front of the vehicle.
Additionally, we introduced the concept of a \emph{region occupancy query} (ROQ), enabling planners to acquire the probability of a region of interest being empty.

Using the FIERY model trained to detect pedestrians on the NuScenes dataset, we demonstrated how to extract these uncertainties from semantic segmentation outputs.
Our experiments highlighted the importance of dedicated pixel-wise and object-wise calibration procedures to ensure the reliability of uncertainty estimates.

In a downstream trajectory planning task, the proposed methods improved planning quality, confirming the practical benefits of well-calibrated uncertainties.
However, this work focused solely on in-distribution uncertainty, and future research should address out-of-distribution uncertainties and dataset shifts.

Additionally, the current approach is limited to binary segmentation.
For real-world applications, extending to multi-class segmentation and calibration methods is essential for obtaining robust uncertainty estimates.
While the core methodology remains similar, multi-class calibration would require an additional probabilistic component to specify the class of the object.
Separate calibration models may also be necessary for different classes to ensure precise uncertainty calibration.

Overall, this work lays the foundation for enhancing uncertainty-aware perception and planning in autonomous driving, with future work in multi-class and out-of-distribution scenarios.

\section*{Acknowledgements}

This work was supported by the Estonian Research Council grant PRG1604, by Bolt Technology O\"U grant LLTAT21278, and by the European Social Fund via the IT Academy programme.

\appendix

%\section{Appendices}

%\appendix
%\section{Appendix}

\section{Code}

%The source code for running experiments and generating visualisations will be made available for the camera-ready version.

The source code for running experiments and generating visualisations is available at \url{https://github.com/markus93/CalPerUn}.

\section{Run time}
\label{sup:runtime}

The run time for finding the four cost maps for three approaches (uncal, pw-cal, ROQ (obj-cal)) described in Section~\ref{subsec:planning} is the following: 
\begin{enumerate}
\item the run time for generating an uncalibrated (uncal) cost map is negligible, 0.0002 seconds on average, as we use the FIERY output segmentation; 
\item the run time for calibrated (pw-cal) segmentation is rapid, 0.001 seconds on average, as the calibration map is precalculated and re-calibrating is fast using isotonic regression; 
\item the run time for finding the cost map of object calibrated uncertainties is relatively slow on average 5.5 seconds (over 4000 frames); however, it could be optimised considerably;
\item the run time for finding the probability of undetected objects in a region costs for 25 trajectories is 0.05 seconds on average. These costs are used in combination with cost maps of object calibrated uncertainties for \emph{obj-cal} approach.
\end{enumerate}

As mentioned previously, the run time for finding the cost maps of object calibrated uncertainties could be decreased a lot, as the bottleneck there is finding object locations as Gaussians with GMMs and converting these back to pixel space after calibration for calculating cost maps. The speed of \emph{obj-cal} approach depends on the number of detected objects.

\section{Additional figures from experiments}
\label{app:extra_exp}

This section has additional figures for Section~\ref{sec:experiments} in the main article.

\begin{figure*}[h]
    \begin{center}
    % -------- Top row (3 plots) --------
    \begin{subfigure}{0.3\textwidth}
        \caption{}
        \includegraphics[width=\textwidth]{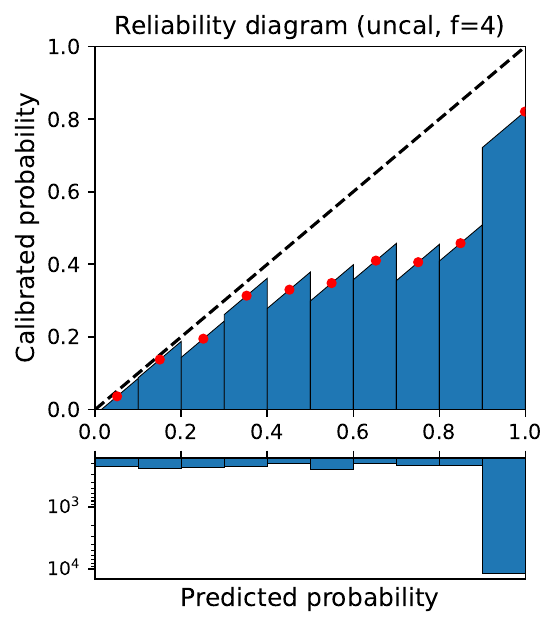}
        \label{fig:rel_diag_uncal_f4}
    \end{subfigure}%
    \hfill
    \begin{subfigure}{0.3\textwidth}
        \caption{}
        \includegraphics[width=\textwidth]{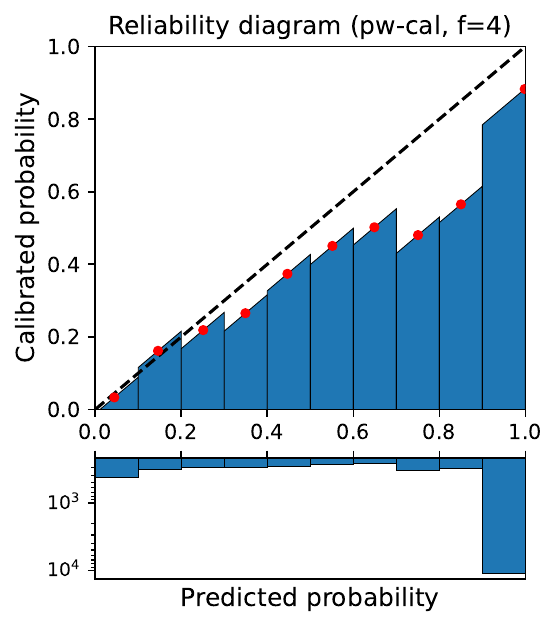}
        \label{fig:rel_diag_gmm_cal_f4}
    \end{subfigure}%
    \hfill
    \begin{subfigure}{0.3\textwidth}
        \caption{}
        \includegraphics[width=\textwidth]{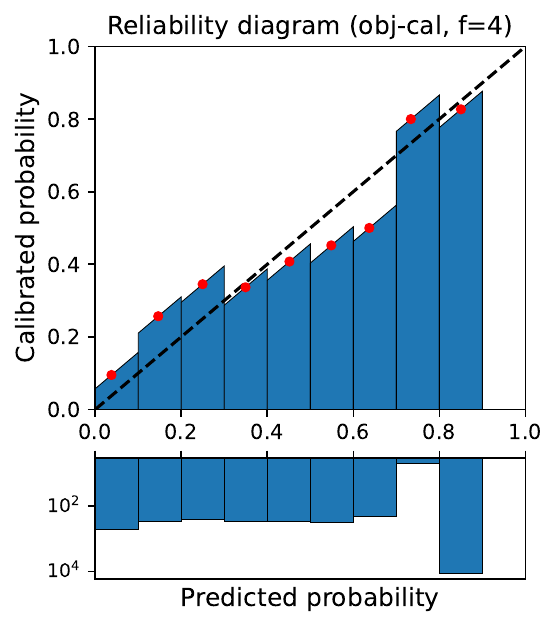}
        \label{fig:rel_diag_obj_cal_f4}
    \end{subfigure}

    \vspace{1em}

    % -------- Bottom row (2 plots) --------
    \begin{subfigure}{0.35\textwidth}
        \caption{}
        \includegraphics[width=\textwidth]{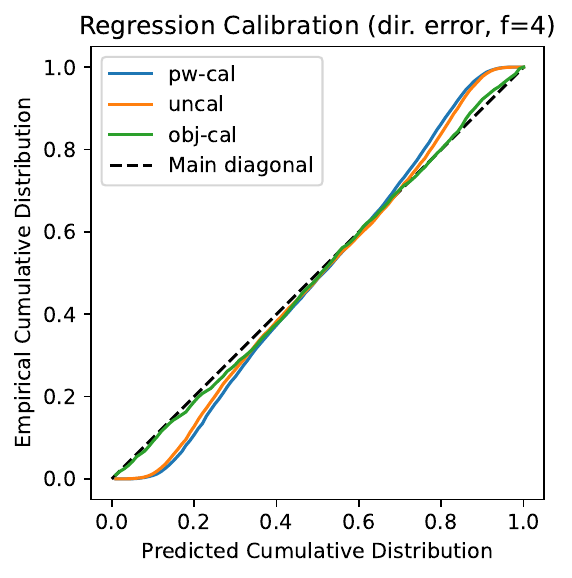}
        \label{fig:cal_location_width_f4}
    \end{subfigure}%
    %\hfill
    \begin{subfigure}{0.35\textwidth}
        \caption{}
        \includegraphics[width=\textwidth]{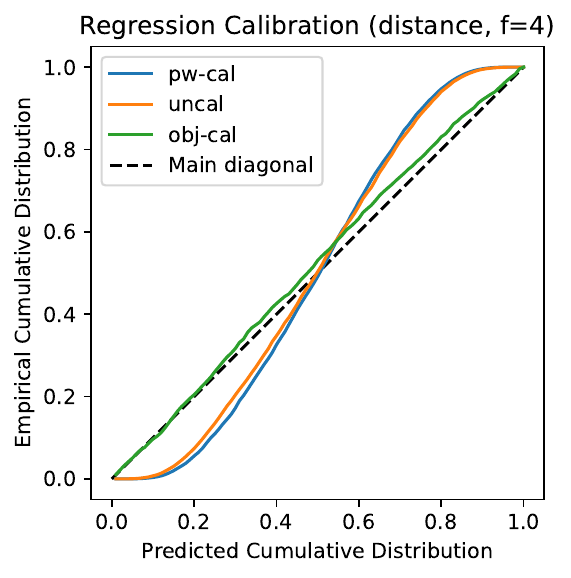}
        \label{fig:cal_location_distance_f4}
    \end{subfigure}

    \caption{
    \modified{\textbf{Top row:} Reliability diagrams of object's presence probabilities:
    (a) uncalibrated,
    (b) pixel-wise calibrated,
    and (c) object-wise calibrated.
    Histograms below the reliability diagrams show the number of elements per bin (log-scaled y-axis). Instances are divided into 10 bins.
    \textbf{Bottom row:} Regression calibration plots~\protect\cite{kuleshov2018accurate} for object's location probabilities:
    (d) direction error and
    (e) distance.
    Black dashed lines indicate perfect calibration.
    Results are for $f=4$ (2 seconds into the future).
    }
    }
    
    \label{fig:calibration_f4_combined}
    \end{center}
\end{figure*}

\modified{Figures~\ref{fig:rel_diag_uncal_f4}--\ref{fig:cal_location_distance_f4} show results for the future timestep ($f=4$).} For uncalibrated and pixel-wise approaches, object presence and location are not well-calibrated. However, object-wise calibration is able to well-calibrate the probabilistic object presence and location uncertainties.

\begin{figure}[h]
    \begin{center}
    \includegraphics[width=0.7\columnwidth]{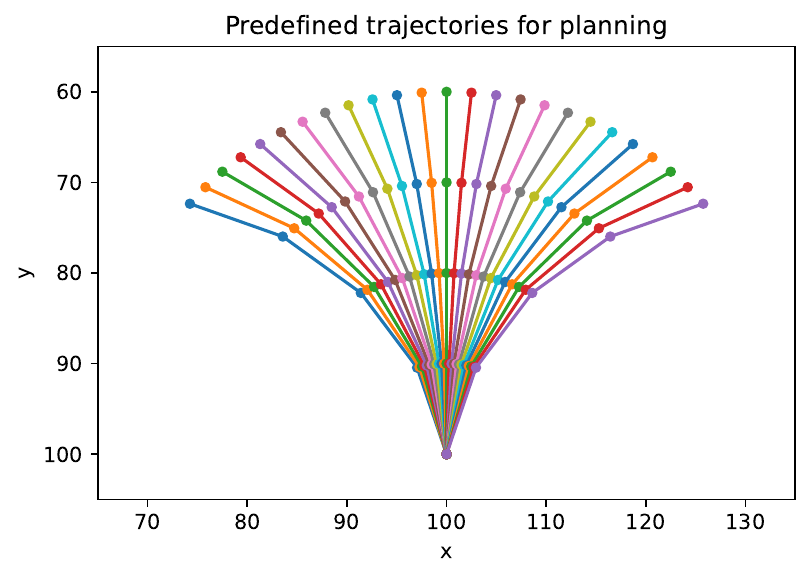}    
    \caption{25 predefined trajectories indicate the route the ego-vehicle will traverse in the next four timesteps (2 seconds).
    }
    \label{fig:predefined_trajectories}
    \end{center}
\end{figure}

Figure~\ref{fig:predefined_trajectories} displays the predefined trajectories used for the downstream planning experiment described in Section~\ref{subsec:planning}.

% Trajectories figure

\FloatBarrier

%\vspace{4cm}

%Figure~\ref{fig:reldiag_pw_cal_es_f4} shows similar results as Figure~\ref{fig:reldiag_pw_cal_es}, only the equal-width binning is replaces with equal-size binning and vice versa. From that we can see that the data is well

%\bibliographystyle{splncs04}
%\bibliography{main}

%%===========================================================================================%%
%% If you are submitting to one of the Nature Portfolio journals, using the eJP submission   %%
%% system, please include the references within the manuscript file itself. You may do this  %%
%% by copying the reference list from your .bbl file, paste it into the main manuscript .tex %%
%% file, and delete the associated \verb+\bibliography+ commands.                            %%
%%===========================================================================================%%
%\clearpage

\section*{References}
\noindent

\bibliography{main}% common bib file
\bibliographystyle{ws-ijufks}

%% if required, the content of .bbl file can be included here once bbl is generated
%%\input sn-article.bbl

\end{document}